%% file: acl2021.tex
\documentclass[11pt,a4paper]{article}

\usepackage[]{emnlp2021}

\usepackage{times}
\usepackage{booktabs}
\usepackage{url}
\usepackage{latexsym}
\usepackage{multirow}
\usepackage{graphicx}
\usepackage{graphics}
\usepackage{chngpage}
\usepackage{caption}
\usepackage{subcaption}
\usepackage{pifont}
\usepackage{microtype}

\title{A Large-Scale Study of Machine Translation in the Turkic Languages}

\author{ \bf
Jamshidbek Mirzakhalov$^{a,b}$,  
Anoop Babu$^{a,b}$, 
Duygu Ataman$^{a,c}$, 
Sherzod Kariev$^{a,b}$, \\
\bf 
Francis Tyers$^{a,d}$,
Otabek Abduraufov$^{a,b}$,
Mammad Hajili$^{a,e}$, 
Sardana Ivanova$^{a,f}$, \\
\bf 
Abror Khaytbaev$^{a,b}$,
Antonio Laverghetta Jr.$^{a,b}$,
Behzodbek Moydinboyev$^{a,b}$,
Esra Onal$^{a,d}$, \\
\bf 
Shaxnoza Pulatova$^{a,g}$,
Ahsan Wahab$^a$, 
Orhan Firat$^{a,h}$,
Sriram Chellappan$^{a,b}$
\\

$^a$Turkic Interlingua, $^b$University of South Florida, $^c$NYU, \\  $^d$Indiana University,  $^e$EPFL,  $^f$University of Helsinki, \\ $^g$Namangan State University,  $^h$Google Research}

\date{}

\begin{document}
\maketitle
\begin{abstract}
Recent advances in neural machine translation (NMT) have pushed the quality of machine translation systems to the point where they are becoming widely adopted to build competitive systems. However, there is still a large number of languages that are yet to reap the benefits of NMT. In this paper, we provide the first large-scale case study of the practical application of MT in the Turkic language family in order to realize the gains of NMT for Turkic languages under high-resource to extremely low-resource scenarios. In addition to presenting an extensive analysis that identifies the bottlenecks towards building competitive systems to ameliorate data scarcity, our study has several key contributions, including, i) a large parallel corpus covering 22 Turkic languages consisting of common public datasets in combination with new datasets of approximately 2 million parallel sentences, ii) bilingual baselines for 26 language pairs, iii) novel high-quality test sets in three different translation domains and iv) human evaluation scores. All models, scripts, and data will be released to the public.\footnote{\url{https://github.com/turkic-interlingua/til-mt}}
 
\end{abstract}

\section{Introduction}

%Language technologies, particularly machine translation (MT), have the potential to break down communication barriers between societies and make information ubiquitously accessible for all. Recent advances in deep learning have drastically increased the potential in building systems that can be used in practice. However, as we scale our systems in terms of size and data resources, we also risk marginalizing many vulnerable populations that lack sufficient data or computational resources, consequently depriving them of the benefits of technological innovations in the space \cite{nekoto2020participatory}. In order to address these problems and illustrate the fall-back of state-of-the-art methods in MT in low-resource languages \cite{joshi2019unsung}, we make the first attempt of studying the practical performance of currently prominent MT methods in a very challenging case of the Turkic language family.%, consisting of a large number of extremely low-resource and morphologically-rich languages. 

Having been studied widely over the last few decades, machine translation (MT) evaluation has traditionally focused on European languages, due to limitations of the available technology as well as resources. Although low-resource MT has recently started to gain more attention and new evaluation benchmarks are becoming available \cite{DBLP:journals/corr/abs-1902-01382, ojha-etal-2020-findings,fraser-2020-findings,ansari2020findings}, there are still a large amount of under-represented languages excluded from MT evaluation. In addition to the cost of preparing such labor-intensive annotations, the lack of training resources also limits the evaluation of MT models in terms of their applicability across a wide range of world languages. On the other hand, many studies have pointed to the limited applicability of prominent methods in MT research including models and evaluation metrics \cite{birch-etal-2008-predicting,stanojevic-etal-2015-results,bugliarello-etal-2020-easier} in translating languages with varying linguistic typology. 

%, consisting of a large number of extremely low-resource and morphologically-rich languages. %Improving evaluation techniques and resources is therefore essential to better understand and develop comprehensive and robust MT models.

\begin{table}[t]
\centering

\resizebox{0.48\textwidth}{!}{%
\begin{tabular}{llrrc}
\toprule
\textbf{Name} & \textbf{Codes} & \textbf{Articles}  & \textbf{Speakers} & \textbf{MT?}\\ 
\midrule
English & en, eng & 6,237,470 & 400M & \ding{51} \\ 
Russian & ru, rus & 1,694,280 & 258M & \ding{51}\\
Turkish & tr, tur & 388,641 & 85.0M & \ding{51}\\ 
Uzbek & uz, uzb & 139,635 & 27.0M & \ding{51}\\
Azerbaijani & az, aze & 177,536 & 23.0M & \ding{51}\\
Kazakh  & kk, kaz & 228,123 & 13.2M & \ding{51}\\
Uyghur  & ug, uig & 4,898 & 10.0M & \ding{51}\\
Turkmen  & tk, tuk & 5,876 & 6.70M & \ding{51} \\
Tatar & tt, tat & 237,332 & 5.20M & \ding{51}\\
Kyrgyz & ky, kir & 80,738 & 4.30M & \ding{51}\\
Bashkir & ba, bak & 55,477 & 1.40M & \ding{51}\\
Chuvash & cv, chv & 45,275 & 1.04M  & \ding{51}\\
Karakalpak & kaa & 1,882 & 583K & \ding{55}\\
Crimean Tatar & crh & 8,633 & 540K & \ding{55}\\
Sakha (Yakut)  & sah & 13,027 & 450K & \ding{51}\\
Kumyk  & kum & --- & 450K & \ding{55}\\
Karachay-Balkar & krc & 2,049 & 310K& \ding{55} \\
Tuvan  & tyv & 3,164 & 280K & \ding{55}\\
Urum & uum & --- & 190K & \ding{55} \\
Gagauz & gag & 2,737 & 148K & \ding{55}\\
Salar  & slr & --- & 70K & \ding{55}\\
Altai & alt &  --- & 56K & \ding{55}\\
Khakas  & kjh & --- & 43K & \ding{55}\\
Shor  & cjs & --- & 3K & \ding{55}\\
\bottomrule
\end{tabular}%
}
\caption{Number of Wikipedia articles for Turkic languages compared to English and Russian along with number of L1 speakers and two- and three-letter language codes. The column \textbf{MT?} indicates if there are currently available online machine translation systems for the language. (K: thousand, M: million.)}
\label{tab:wikipedia}
\end{table}

In order to extend the evaluation of the state-of-the-art methods in MT \cite{joshi2019unsung} and ultimately aid in designing methods with wider range of applicability, in this paper, we present a large-scale case study of MT methods in a very challenging case of the Turkic language family.
The Turkic language family consists of around 35 languages spoken by communities across Eurasia by around 200 million people. Of this number, around 20 are official languages of a state, or sub-national entity, with the remaining being minority languages. The languages are distinct in their highly complex use of morphology, and thus create extremely sparse vocabularies, presenting a challenging case of evaluation of MT systems \cite{tantug2008bleu+} and generally leading to worse performance in n-gram language models \cite{bender2011achieving, tsarfaty-etal-2020-spmrl}.
Table~\ref{tab:wikipedia} presents the amount of resources and the number of speakers in Turkic languages\footnote{https://www.ethnologue.com/} which aids our analysis on the feasibility in crowdsourcing, based on the approach of \citet{moshagen:14}.

%In order to assess the amount of resources and the potential for crowdsourcing in Turkic languages, we conduct an analysis as proposed by \citet{moshagen:14}, 
%based on the assumption that larger the number of speakers and articles in Wikipedia, the more likely it would be that crowdsourcing is feasible. 
%presented in Table~\ref{tab:wikipedia} highlighting the potential in crowdsourcing.% presents the list of Turkic languages with the number of L1 speakers, the number of Wikipedia articles available in each language and whether their MT is supported through any online services.
%To give an idea of the potential for crowdsourcing for the various languages, we follow \cite{moshagen:14} and use Wikipedia as a measure of practicality of crowdsourcing. The larger the number of speakers and articles in the Wikipedia, the more likely crowdsourcing is to be feasible. Table~\ref{tab:wikipedia} presents the list of Turkic languages with the number of speakers and the number of Wikipedia articles available in each language and whether their translation is supported through any online services.
%\TODO{this is the section that probably needs some story change}
Our study includes the preparation of novel public resources covering many languages in the Turkic family, most of which included for the first time in parallel corpora. We also present new benchmarks for MT which could be used for assessing different factors determining the limits of MT methods in various languages, such as data size, evaluation metrics, translation domain, linguistic typology, relatedness, and the writing system. We test the use of our resources in MT and present the first evaluation results for many Turkic languages.
Our novel resources consist of \textit{i)} a large-scale multi-centric parallel corpus of 75M+ sentence pairs in 22 Turkic languages and their translations into English, Russian, as well as in-family languages, covering over 400 translation directions,
\textit{ii)} 3 new test sets for each translation direction curated from our corpus in 3 different translation domains, \textit{iii)} bilingual baselines in 26 different language pairs. Our baselines are evaluated using automatic metrics as well as human assessments against commercial or open-source systems where applicable. We release our parallel corpora, test sets, and baseline systems publicly to encourage future research in Turkic languages.

% In addition to conventional machine translation evaluation, our corpus provides an extensive evaluation benchmark for studying many of the recently addressed problems in machine translation and multi-lingual language modeling, such as multi-centricness (a new word?), language relatedness, amount of training data and the effects of morphological typology and scripts in language transfer. 

% TODO: Write about group research structure and low-resource mt being similar to Masakhane paper

%The following instructions are directed to authors of papers submitted to ACL-IJCNLP 2021 or accepted for publication in its proceedings.
%All authors are required to adhere to these specifications.
%Authors are required to provide a Portable Document Format (PDF) version of their papers.
%\textbf{The proceedings are designed for printing on A4 paper.}

\section{Turkic Languages \& MT}

This section gives a brief overview of Turkic languages from a linguistic perspective as well as highlights the previous work on MT of these languages. In our study, we include %Our corpus contains parallel sentences for 
22 Turkic languages: Altai, Azerbaijani, Bashkir, Crimean Tatar, Chuvash, Gagauz, Karachay-Balkar, Karakalpak, Khakas, Kazakh, Kumyk, Kyrgyz, Sakha, Salar, Shor, Turkmen, Turkish, Tatar, Tuvan, Uyghur, Urum, and Uzbek. There are several other widely spoken languages that were left out from %the corpus 
our study such as Nogai, Khorasani Turkic, Qashqai, and Khalaj, due to the lack of any available parallel corpora. Future work will focus on extending the corpus to these languages as well. 
% INCLUDE EXACTLY WHICH LANGUAGES ARE IN, WHICH ARE OUT AND WHY.

% OUT: Nogai, Khorasani, Khalaj, Chulym, Dolgan, ...

\subsection{Linguistic Typology}
The Turkic languages are spoken in a wide area that stretches from south-eastern Europe to north-eastern Asia. The languages are of the agglutinative morphological type and uniformly have Subject-Object-Verb main constituent order. 

Nominal morphology is highly similar between the languages, with all of them exhibiting inflection for number, possession, and case. There are a variable number of cases, but the six-core cases of nominative, genitive, accusative, dative, locative, and ablative are extant in the vast majority of languages. As part of the nominal inflectional system, the languages also have a derivational process whereby locatives and genitives can be pronominalized and constitute full noun phrases in their own right.
Verbal inflection, on the other hand, is more heterogeneous between the languages with each language having a variety of strategies for encoding tense, aspect, voice, modality, and evidentiality. One common feature however is that each of the languages has an extensive system of non-finite forms: verbal adjectives, verbal nouns, and verbal adverbs. These are full clauses that can be used as either modifiers (in the case of verbal adjectives and verbal adverbs) or heads (in the case of verbal nouns). Many of the languages also have constructions consisting of a non-finite verbal form and an auxiliary verb which constitute a single predicate, with the auxiliary verb giving extra information about tense or mood \cite{johanson2015turkic}.

The modern Turkic languages are written in a variety of scripts, with Latin, Cyrillic, and Perso-Arabic being most common. Many of the languages have been written in several writing systems over the past century, making collecting texts more problematic. For example, we can find instances where the same language have texts that are written in Perso-Arabic before the 1920s, in Latin until the 1930s, in Cyrillic until the 1990s, and then in Latin again \cite{rona-tas:15}.  In addition, many languages have gone through several orthographic norms based on the same script, and some languages are currently written in different scripts depending on which country the speakers are in. This orthographic diversity makes collecting and collating text resources difficult, as many texts may be available only in a previously-used orthography and conversion between orthographic systems is never deterministic owing to the large number of loan words in many texts.

\subsection{MT of Turkic Languages}

The need for more comprehensive and diverse multilingual parallel corpora has sped up the creation of such large-scale resources for many language families and linguistic regions \cite{koehn2005europarl, choudhary2011creating, post2012constructing, nomoto2018tufs, espla2019paracrawl, nekoto2020participatory}. \citet{tiedemann2020tatoeba} released a large-scale corpus for over 500 languages covering thousands of translation directions. The corpus includes 14 Turkic languages and provides bilingual baselines for all translation directions present in the corpus. However, the varying and limited size of the test sets %(as low as 1 sentence) 
does not allow for the extensive analysis and comparisons between different model artifacts, linguistic features, and translation domains.  
%There have been several studies in building MT systems and parallel corpora for Turkic languages. 
\citet{khusainov2020first} collected a large-scale Russian-Turkic parallel corpus for 6 language pairs and reports bilingual baselines using a number of NMT-based approaches, although the dataset, test sets, and the models are not released to the public which limits its use to serve as a comparable benchmark. \citet{alkim2019machine} introduces a rule-based MT framework for Turkic languages and demonstrates the performance with 4 language pairs.  \citet{apertium:2019} demonstrates several rule-based MT systems built for Turkic languages which are available through the Apertium\footnote{\href{https://www.apertium.org/}{https://www.apertium.org/}} website. %Unfortunately, the scalability and extensibility of rule-based MT systems still remain an open question. 

For individual languages in our corpus, there are several proposed MT systems and linguistic resources: Azerbaijani \cite{hamzaoglu1993machine, fatullayev2008dilmanc}, Bashkir \cite {tyers2012prototype}, Crimean Tatar \cite {gokirmak2019machine, altintacs2001turkish}, Karakalpak \cite {kadirov2015algorithm}, Kazakh \cite{assylbekov2014initial, sundetova2015free, littell2019multi, briakou2019university, tukeyev2019neural}, Kyrgyz \cite{ccetin7assisting}, Sakha \cite{ivanova-etal-2019-tools}, Turkmen \cite {Tantug2007MachineTB}, Turkish \cite{turhan1997english,el2006initial,bisazza2009morphological,tantuug2011turkmenceden, ataman2017linguistically}, Tatar \cite{salimzyanov2013free, khusainov2018building, valeev2019application, gokirmak2019machine}, Tuvan \cite{killackey2013statistical}, Uyghur \cite{mahsut2004experiment, nimaiti2014rule,  song2015construction, wang2020njunlp}, Uzbek \cite {axmedova2019algorithm}. Yet, to our knowledge, there has not been a study that covers Turkic languages in such a large extent as ours, both in terms of multi-lingual parallel corpora and benchmarks including multi-way comparable test sets in all languages.

\input{languages}

\section{TIL Corpus}

Our parallel corpus is collected through unifying publicly available datasets and additional parallel data we prepare by crawling public domain resources. Table~\ref{tab:data} shows the total amount of sentences in that particular language across the corpus along with number of sentences that are newly introduced (previously unavailable). 
This section describes the details of our data collection process. 

\subsection{Public Datasets}
In our corpus we include the following public data sets:
\begin{itemize}
  \item The Tatoeba corpus \cite{tiedemann2020tatoeba} provides training and test sets for over 500 languages and thousands of translation pairs. It uses the latest version of OPUS\footnote{\href{http://opus.nlpl.eu/}{http://opus.nlpl.eu/}} \cite{tiedemann2004opus} as training sets and use parallel sentences from the Tatoeba project for testing. Tatoeba consists of 58 language pairs of interest. For the purposes of our corpus, we merge the training, development, and test sets into a single set for all available languages.  
    \item JW300 \cite{agic-vulic-2019-jw300} is a public dataset available for download through OPUS. Although most of the parallel data in JW300 was provided through the Tatoeba corpus, we have identified several pairs that were missing in Tatoeba but present in JW300. To avoid further data loss, we have obtained the JW300 dataset directly from OPUS and deduplicated it against the Tatoeba corpus. This dataset provided data for 59 language pairs of interest and resulted in 5.2 million parallel sentences. % before deduplication. 
    \item GoURMET\footnote{\href{https://gourmet-project.eu/}{https://gourmet-project.eu/}} is another dataset available through OPUS and provides parallel sentences for 7 language pairs including English-Turkish and English-Kyrgyz. They are not available in Tatoeba due to a more recent release. English-Kyrgyz consists of 14.5 thousand sentence pairs while English-Turkish contains 1.3 million. 
\end{itemize}

In addition to this, with the permission from the owners, we include privately owned corpora for English-Azerbaijani\footnote{\href{https://github.com/derintelligence/en-az-parallel-corpus}{https://github.com/derintelligence/en-az-parallel-corpus/}} containing data from news articles, English-Uzbek\footnote{\href{https://uz.khanacademy.org/}{https://uz.khanacademy.org/}} containing data from KhanAcademy website localization, and Bashkir-Russian\footnote{\href{https://github.com/AigizK/bashkort-parallel-corpora
}{https://github.com/AigizK/bashkort-parallel-corpora
}} having a mix of data from news articles and literary works.

\subsection{Data Crawling}

We obtained additional parallel data from a few different public domain websites that contain a large amount of text translated into many different languages. One of these includes \textit{TED Talks},\footnote{\href{https://www.ted.com/talks}{https://www.ted.com/talks}} which contains talks across various domains that are translated by volunteers. \citet{qi2018pretrained} compiled a dataset for 60 languages, however, only a few Turkic languages were available at their time of curation. We have compiled an updated version of this dataset and obtained sentence pairs for 8 Turkic languages. \textit{Bible.is}\footnote{\href{http://www.bible.is}{http://www.bible.is}} is another website that contains an extensive list of languages into which religious texts and books are translated. 19 out of 22 Turkic languages were covered in this source with an average of approximately 8,000 sentence pairs for each translation direction. Additionally, we have crawled other public websites, online dictionaries, and resources with parallel data that were identified by native speakers of these languages. The full list of online resources we used in our crawling is given in the Appendices.

% % \input{datatable.tex}

% \begin{table}[]
% \centering
% \resizebox{6.25cm}{!}{%
% \begin{tabular}{lrr}
% \toprule
% \textbf{Name} & \textbf{Avg. sents.}  & \textbf{Paired with}\\
% \midrule
% Turkish         & 3.0M   & 20 \\
% Kazakh          & 286.6K & 19 \\
% Uzbek           & 142.1K & 20 \\
% Azerbaijani     & 140.0K & 20 \\
% Tatar           & 126.8K & 20 \\
% Kyrgyz          & 119.6K & 20 \\
% Turkmen         & 64.9K  & 20 \\
% Chuvash         & 49.1K  & 20 \\
% Bashkir         & 24.2K  & 20 \\
% Uyghur          & 18.5K  & 19 \\
% Tuvan           & 16.3K  & 10 \\
% Altai           & 8.4K   & 20 \\
% Crimean Tatar   & 8.1K   & 19 \\
% Sakha           & 7.0K   & 19 \\
% Karakalpak      & 7.0K   & 19 \\
% Khakas          & 6.9K   & 19 \\
% Karachay-Balkar & 6.9K   & 19 \\
% Kumyk           & 6.7K   & 19 \\
% Gagauz          & 6.7K   & 19 \\
% Shor            & 2.4K   & 1  \\
% Salar           & 578    & 1  \\
% Urum            & 507    & 1 \\
% \bottomrule
% \end{tabular}%
% }
% \caption{Corpus statistics. Column 2 represents the average number of parallel sentences. (K: thousand, M: million.)}
% \label{tab:Data}
% \end{table}

\subsection{Data Alignment}
\label{sec:preprocessing}

All crawled documents are aligned using Hunalign \citep{varga:05}, with a threshold of either 0.2 or 0.4 depending on the availability of a native speaker for the language. When crawling pre-aligned sources such as TED Talks, we noticed serious alignment issues with certain Turkic languages, especially when the source and target differ greatly in size. In these cases, we split both sides into sentences using NLTK sentence tokenizer\footnote{\href{http://www.nltk.org/api/nltk.tokenize.html}{http://www.nltk.org/api/nltk.tokenize.html}} and realign using the Hunalign tool.   
Specifically for the Bible dataset, all the data has been aligned at the verse level first, then split into sentence-level bitexts whenever possible. This results in parallel texts that are relatively longer while ensuring higher quality alignments.
%1-3 sentences long, but this approach would ensure the high quality of alignment at the cost of longer sequences.

\subsection{Data Preprocessing}

Many of the languages in our dataset are written using multiple scripts, which creates consistency problems for building MT systems. %When NMT systems are built trained with languages that has multiple scripts, the script of the output sequence becomes almost impossible to predict which makes the evaluation unnecessarily hard. 
Therefore, we transliterate three of the languages in our dataset that have a high mix of multiple scripts. Namely, we transliterate Uzbek into a Latin script, while all Karakalpak text is converted into Cyrillic. Although the performance of transliteration tools (Uzbek\footnote{\href{https://github.com/kodchi/uzbek-transliterator}{https://github.com/kodchi/uzbek-transliterator}} and Karakalpak\footnote{\href{http://www.transliteration.kpr.eu/kaa/}{http://www.transliteration.kpr.eu/kaa/}}) were not strictly evaluated, the tools we have used were recommended and widely adopted by the native speakers of the languages. Once we combine the entire corpus data, we deduplicate the sentences in each language pair. 

\section{Bilingual Baselines}

We train bilingual baselines for 26 language pairs in three different resource categories: high (>5M), medium (100K-5M) and low (<100K). The choice of pairs to train was based on multiple factors such as the availability of test sets, native speakers (for human evaluation), and other comparable MT systems.

% \usepackage{graphicx}

\subsection{Model Details}

All models are Transformers \cite{vaswani2017attention} (\textit{transformer-base}) whose exact configuration depends on the amount of data available for training. Models for low-resource pairs use 256-dimensional embeddings and hidden layers. Models for mid-resource pairs use 512-dimensonal embeddings and hidden layers. The models for high-resource pairs use the same 512-dimensonal embedding and hidden layer sizes for the encoder, but for the decoder both dimensions are increased to 1024. All models are trained with the Adam optimizer \cite{kingma2015adam} over cross-entropy loss with a maximum learning rate of $3*10^{-4}$ and a minimum of $1*10^{-8}$, which warms up for the first 4800 training steps and then decays after reaching the maximum. We use a training batch size of 4096. We use perplexity as our early stopping metric with a patience of 5 epochs. We set a dropout \cite{srivastava2014dropout} probability of 0.3 in both the encoder and the decoder. 
We apply a byte pair encoding (BPE) \cite{sennrich2015neural, dong-etal-2015-multi} with a joint vocabulary size of 4K and 32K for low- and mid/high-resource scenarios respectively. 

All models use the Joey NMT \cite{kreutzer-etal-2019-joey} implementation and apex\footnote{\href{https://github.com/NVIDIA/apex}{https://github.com/NVIDIA/apex}} where possible to speed up training. Models were trained on preemptible GPUs freely available on Google Colab.\footnote{\href{https://colab.research.google.com/}{https://colab.research.google.com/}} 

\subsection{Test Sets}
High-quality and diverse test sets are essential in evaluating the strength and weaknesses of MT systems. We curate 3 test sets covering 3 translation domains: religious (Bible), conversational (TED Talks), and news (X-WMT). 

\paragraph{Bible}
\label{evaluation:bible}
dataset is the main source that exists across almost all of the 24 language pairs that are included in our corpus. From this dataset, around 400 to 800 most commonly present sentences for every language pair were separated to create a test set. This allowed having a test set comparable in all language pairs, which we find essential for a controlled evaluation and believe would be a useful resource in future studies involving multilingual models.   

\paragraph{TED Talks}
\label{evaluation:ted}
is another resource we use for collecting sentences across multiple languages to create a language-wise comparable test set in the conversational domain. This allows our approach to be comparable also across different domains. After deduplication, 3000-5000 sentences per language pair are picked as a part of our TED Talks test set.

\paragraph{X-WMT}
\label{evaluation:x-wmt}
is our test set in the news domain based on the professionally translated test sets in English-Russian from the WMT 2020 Shared Task \citep{mathur-etal-2020-results}. This set contains approximately 1,000 sentences curated both from English and Russian-centric news sources. Through the engagement of native speakers and professional translators\footnote{A total of 22 volunteer translators}, we partially translate this test set into 8 Turkic languages (Bashkir, Uzbek, Turkish, Kazakh, Kyrgyz, Azerbaijani, Karakalpak, and Sakha). 

\begin{table}[]
\centering
\resizebox{\columnwidth}{!}{%
\begin{tabular}{lrrrrrrrrrr}
\toprule
 & \textbf{en} & \textbf{ru} & \textbf{ba} & \textbf{tr} & \textbf{uz} & \textbf{ky} & \textbf{kk} & \textbf{az} & \textbf{sah} & \textbf{kaa} \\ 
 \midrule
\textbf{en}  & ---    &     &   &  &  &  &  &  &  &  \\ 
\textbf{ru}  & \textbf{1000}  &  ---   &   &  &  &  &  &  &  &  \\  
\textbf{ba}  & 1000  &  \textbf{1000}   & --- &   &  &  &  &  &  &  \\  
\textbf{tr}  & \textbf{800} & 800 & 800  &  --- &  &  &  &  &  &  \\  
\textbf{uz}  & \textbf{900} & 900 & 900 & 600 &  --- &  &  &  &  &  \\  
\textbf{ky}  & 500 & \textbf{500} & 500 & 400 & 500 & ---&  &  &  &  \\ 
\textbf{kk}  & 700 & 700 & 700 & 500 & \textbf{700} & 500 & ---&  & & \\  
\textbf{az}  & \textbf{600} & 600 &  600 & 500  & 600  & 500  & 500 & --- &  &   \\  
\textbf{sah} & 300 & \textbf{300} & 300 & 300 & 300 & 300 & 300 & 300 &  ---&   \\ 
\textbf{kaa} & 300 & 300 &  300  & \textbf{300}   &  300  & 300 & 300 & 300 & 300 & ---\\
\bottomrule
\end{tabular}%
}
\caption{X-WMT test sets. Bolded entries indicate the original translation direction.}
\label{tab:wmt}
\end{table}

Table~\ref{tab:wmt} highlights the currently available test set directions.
Bolded entries in the table indicate the original direction of the translation. While Bashkir and Sakha have been translated by professional translators, other languages have been translated and validated (by another person) by proficient bilingual speakers of both the source and target language. The curation of this test set is an ongoing and growing effort currently covering 88 language directions.

\input{bilingual}

\section{Evaluation}
\label{evaluation}

Automatic evaluation metrics are very commonplace in MT research, and there has been a recent line of work exploring better metrics that capture translation quality beyond the syntactic and lexical features \cite{zhang2019bertscore, sellam2020bleurt, rei2020comet}.
Methods relying on contextual embeddings to capture the semantic similarity between the hypothesis and references fall short in terms of their language coverage. This is largely due to the pretraining of these evaluation models that require a significant of monolingual data which most of the low-resource languages lack. In this study, we evaluate our systems using both automatic metrics and human evaluation of translations.

\subsection{Automatic Metrics for MT}

We employ two widely adopted metrics: BLEU~\cite{papineni2002bleu} and ChrF~\cite{popovic-2015-chrf}. BLEU utilizes modified \(n\)-gram precision where the consecutive \(n\)-grams of the system translation are compared with the consecutive \(n\)-grams of the reference translation. We use the standard SacreBLEU implementation \cite{post-2018-call}. ChrF applies the same method at the level of character \(n\)-gram and we use the original implementation from the paper as provided through NLTK library.\footnote{\url{https://github.com/m-popovic/chrF}}

\subsection{Human Evaluation}
\label{human-evaluation}

To perform a more holistic analysis of MT systems, it is critical to involve native speakers in the evaluation process. We conducted a human evaluation campaign using a randomly sampled subset of 250 sentences from X-WMT or Bible (whenever X-WMT was not available) for evaluating the outputs of 14 bilingual baseline models. Our assessment is based on Direct Assessment (DA) test \cite{niessen2000evaluation, papineni2002bleu, doddington2002automatic}, where annotators were asked to rate a translation according to \textit{adequacy} and \textit{fluency} on a 5 point Likert scale. All participants of the study were bilingual speakers of the source and target language. To better understand the importance of directionality (e.g. English-X vs X-English) and avoid variance in scores, we ensure that both directions of the same pair are evaluated by the same annotator (whenever possible). While reporting, we average the scores for each pair but report \textit{adequacy} and \textit{fluency} separately. Adequacy is defined as how much information is preserved in the translation. A score of 1 would mean that the translation is meaningless and has no correlation with the target sentence. A score of 5 would mean the translation retains all of the information. Fluency is defined as how grammatically, syntactically, and stylistically correct the translation is. A score of 1 would mean the sentence makes no sense grammatically or syntactically. A score of 5 would mean the sentence is perfectly correct.   

\section{Results \& Discussion}

\begin{table}[]
    \centering
    \scalebox{0.96}{
    \begin{tabular}{lrrrr}
    \toprule
               & \multicolumn{2}{c}{\textbf{Adequacy}} & \multicolumn{2}{c}{\textbf{Fluency}} \\
               \midrule
               & \textbf{BLEU} & \textbf{ChrF} & \textbf{BLEU} & \textbf{ChrF} \\\cline{2-5}
      \textbf{Turkic}   &  0.62     & {\bf 0.71}    &  0.75    &  0.67    \\
      \textbf{Non-Turkic} &  {\bf 0.75}     & 0.68 & \textbf{0.83}        &  {\bf 0.86}    \\
      \bottomrule
    \end{tabular}
    }
    \caption{Correlation between scores from human evaluation and automatic metrics for translating into Turkic and non-Turkic. Correlation is measured using Pearson's \(r\).}
    \label{tab:metric-correlation}
\end{table}

The upper section of Table~\ref{tab:comb-bilingual} highlights the bilingual baselines for high-resource pairs and their evaluation scores in the three domains. Despite the large training size, both models perform relatively modestly on the Bible and TED Talks with the \textit{en-tr} model slightly better than \textit{ru-tr}. Our hypothesis is that the domain of the Bible test set is far from the rest of the training set for both pairs, as most of the training data for Turkish comes from OpenSubtitles.\footnote{\href{http://www.opensubtitles.org/}{http://www.opensubtitles.org/}} Another likely bottleneck is the suboptimal model size and hyperparameters, which were not tuned  due to limited computational resources.

Baseline results for the mid- and low-resource pairs are in the lower part of Table ~\ref{tab:comb-bilingual}. While there are a lot of fluctuations in the results, it is important to note the large disparities in BLEU scores between models when translated in and out of non-Turkic languages. However, these differences are not as prominent when evaluated using ChrF, which is a character-level metric. 
This can partially be attributed to the complex morphology of Turkic languages which penalizes lexical mispredictions at a much higher rate than in English for example \cite{tantug2008bleu+}. This in return would lead to lower BLEU scores. To examine this phenomena in more detail, we compare the results of X-WMT against human evaluations for the translations these models produced in Section ~\ref{human2bleu}.
%, which measures the consecutive \(n\)-gram precision over space separated tokens between the hypothesis and target sentence.

Another notable aspect is the importance of scripts in the performance of the models. Language pairs with more than one script consistently under-perform (both in automatic and human evaluations) the ones where both the source and target language use the same script. In fact, the best 6 models on the X-WMT test sets all have Latin scripts in both the source and target language. A suboptimal performance in the face of a script disparity is a known phenomenon \cite{anastasopoulos2019pushing, murikinati2020transliteration, aji2020neural, amrhein2020romanization}, where techniques such as transliteration show to improve performance.  This is mostly attributable to model’s inability to represent
both languages in a shared space effectively when they do not share the same script, which can be damaging for the downstream performance. 
\begin{figure*}[htbp]
     \centering
     \begin{subfigure}[c]{0.45\textwidth}
        \centering
        \includegraphics[width=1\textwidth]{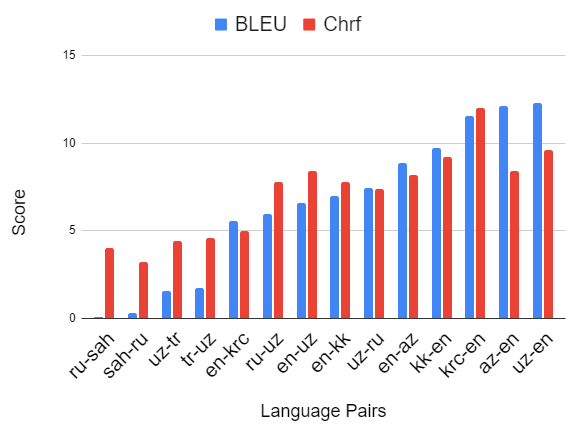}
        \caption{BLEU and ChrF scores for select pairs. Note: ChrF scores were multiplied by 20 for better visibility.}
        \label{fig:bleu_chrf}
     \end{subfigure}
     \hspace*{\fill}%
     \begin{subfigure}[c]{0.45\textwidth}
        \centering
        \includegraphics[width=1\textwidth]{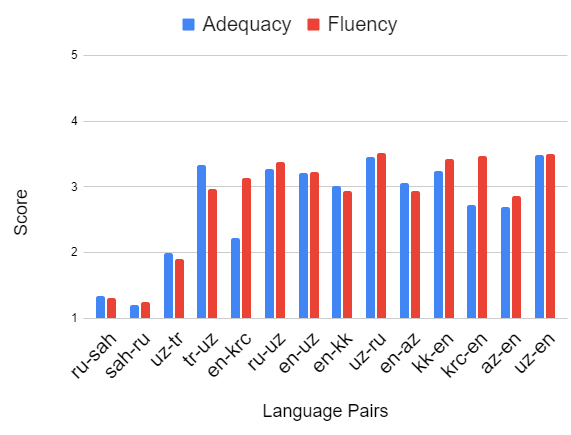}
        \caption{Adequacy and Fluency scores (1-5) obtained from human evaluations.}
        \label{fig:adequacy_fluency}
     \end{subfigure}
     \caption{Comparison between BLEU/ChrF scores and Adequacy/Fluency scores. Best viewed in color.}
     \label{fig:bleu_adflu}
\end{figure*}

% \begin{figure}[h]
%      \centering
%      \includegraphics[width=1\linewidth]{acl-ijcnlp2021-templates/graphs/bleu and chrf.png}
%      \caption{Adequacy and Fluency scores (1-5) obtained from human evaluations.}
%      \label{fig:adequacy_fluency}
     
% \end{figure}

\subsection{Comparing Human Evaluations to BLEU}
\label{human2bleu}
Using the Direct Assessment (DA) surveys described in Section ~\ref{human-evaluation}, we obtain average scores of \textit{adequacy} and \textit{fluency} for almost all baseline models. Figures \ref{fig:bleu_adflu} show the scores for BLEU/ChrF and \textit{adequacy}/\textit{fluency} respectively. Comparing the scores from native speakers of these languages, it is quite evident that the disparities in BLEU scores between two translation directions are exaggerated and, even misleading (e.g. \textit{en-az} vs \textit{az-en}). Results in the human evaluations for mid-resource pairs seem a lot more closely clustered than in the BLEU/ChrF figure. These results further emphasize the pitfalls of automatic metrics of MT evaluation and emphasize the role of native speakers in the MT process.

%non-turkic:
%BLEU-Adequacy	0.7596909466
%BLEU-Fluency	0.8342967835
%Chrf-Adequacy	0.6825922632
%Chrf-Fluency	0.8561491444
%Turkic:
%BLEU-Adequacy	0.6239063607
%BLEU-Fluency	0.7488166999
%Chrf-Adequacy	0.713378007
%Chrf-Fluency	0.6765956006

\subsection{Turkic Languages on the target side}

Even though BLEU scores do not offer a holistic way to compare two MT systems, they are effective in telling which system performs better. As seen clearly from the results in Table ~\ref{tab:comb-bilingual}, the performance of the baseline system as measured by the BLEU metric when translating into a Turkic language from English is substantially worse than when translating into English from a Turkic language. Translating into the Turkic language is typically twice as bad in terms of BLEU as translating from the Turkic language. The reliability of the BLEU score also decreases especially in the case of translating into morphologically-rich languages, which has indeed been shown to correlate poorly with human judgments in Turkic languages \cite{ma2018results,ma2019results}. Table ~\ref{tab:metric-correlation} shows the  correlation between BLEU/Chrf and \textit{adequacy}/\textit{fluency} scores. BLEU seems to correlate with \textit{adequacy}/\textit{fluency} a lot better when the target side is a non-Turkic language, which emphasizes our earlier points regarding the language morphology. ChrF's correlation to \textit{adequacy} scores is about the same regardless of the target language.

\subsection{Comparison to Existing Systems}

Table ~\ref{tab:bilingual--hr--comparisons} compares our baselines to three commercial/open-source MT systems: Google Translate,\footnote{\url{https://translate.google.com/}} Yandex Translate,\footnote{\url{https://translate.yandex.com/}} and Apertium \cite{forcada2011apertium}. Google Translate results are significantly higher than our baselines and other MT systems. There are quite a few reasons for the score disparities. First, commercial systems have access to more data for training and possibly also include the public data we exclude from our test sets.
%Google Translate utilizes a lot more data for training and benefit from cross-lingual transfer learning in a multilingual setup. Second, there is no guarantee that the original test sets were not part of the training data for the system which would make the comparisons less meaningful. Last but not least, 
Moreover, several test-set translators used Google Translate to do the translations and performed post-edits afterwards (e.g. \textit{en-uz}) which creates a bias  favoring sentences generated by Google's service. A safer comparison of the baselines %would be with 
is achieved with Yandex Translate, which despite the lower performance also supports more Turkic languages (8 in Google and 9 in Yandex).  However, it is important to note that their API yielded worse results than their web interface. 
Apertium is a rule-based MT framework that supports several Turkic-Turkic pairs and we include the results whenever one is available. For those pairs, the results are comparable with our baselines and Yandex Translate.

\section{Conclusion \& Future Work}

% multilingual model

% include more languages

% maybe reword this?
In this paper, we introduce a large parallel corpus covering 22 Turkic languages along with in-domain and out-of-domain evaluation sets. We also train the first baseline models for several language pairs and take the initial steps to address the challenges associated with machine translation in the Turkic languages. This study was carried out as in a participatory research setting by a diverse community of researchers, engineers, language specialists, and native speakers of Turkic languages. Future work will focus on studies of methods for effective cross-lingual transfer, extending of the coverage of the corpus to more languages and domains, and increasing the size of the test sets to provide more comprehensive benchmarks. 

\section*{Acknowledgements}
This project received support from the Google AI Academic Research Awards and the Swiss National Science Foundation (MUTAMUR; no.\ 176727). We thank all of the members and partners of the Turkic Interlingua (TIL) community for their contributions to the project. Namely, we would like to thank our dedicated translators and annotators: Nurlan Maharramli, Leyla Baghirli, Ipek Baris, Aigiz Kunafin, Aydos Muxammadiyarov, Ziyodabonu Qobiljon qizi, Alperen Cantez, Doniyorbek Rafikjonov, Mukhammadbektosh Khaydarov, Madina Zokirjonova, Erkinbek Vokhabov, Mohiyaxon Uzoqova, Petr Patrov, Abilxayr Zholdybai, and Akylbek Khamitov. We also acknowledge and appreciate significant dataset contributions from Rasul Karimov, Iskandar Mamasoliev, Khan Academy O'zbek, and the Foundation for the Preservation and Development of the Bashkir Language. Furthermore, we would like thank Dr. John Licato, Dr. Jonathan Washington and Animesh Nighojkar for their valuable feedback throughout the project.

\bibliographystyle{acl_natbib}
\bibliography{anthology,acl2021,mirzakhalov,Antonio, lit_review}

\appendix

\section{Overall corpus statistics}

Table~\ref{tab:til-data} lists the training size (in sentences) for each language direction. It is important to note that the corpus is continuously growing and improving, so this version of the corpus was used for the bilingual baselines and human evaluations in this paper.

\input{datatable}

\section{Dataset Sources}
Our parallel corpus is a combination of public resources and individual/group contributions. We list the sources for all the resources and websites used in curating our corpus in Table~\ref{tab:sources}. More recent information on the licences and reuse of the corpus can be found in the Github repository\footnote{\url{https://github.com/turkic-interlingua/til-mt/tree/master/replicate_results/corpus_paper}}. 

\input{sources}

\end{document}

%% file: languages.tex
\begin{table*}[t]

\centering
\begin{tabular}{lrllr}
\hline
\multicolumn{1}{l}{\textbf{Language}} & \multicolumn{1}{r}{\textbf{Data}} & \multicolumn{1}{l}{\textbf{Script}} & \multicolumn{1}{l}{\textbf{Category}} & \multicolumn{1}{r}{\textbf{New Data}} \\ \hline
Turkish                           & 52.6M                             & Latin                               & The Underdogs (4)                          & 755.9K                                \\
Kazakh                            & 5.3M                              & Arabic, Cyrillic, Latin             & The Rising Star (3)                        & 201.9K                                \\
Uzbek                             & 2.9M                              & Arabic, Cyrillic, Latin             & The Rising Star (3)                        & 1.7M                                  \\
Azerbaijani                       & 2.2M                              & Arabic, Cyrillic, Latin             & The Scraping-Bys (1)                       & 284.8K                                \\
Tatar                             & 1.8M                              & Arabic, Cyrillic                    & The Scraping-Bys (1)                       & 192.0K                                \\
Kyrgyz                           & 1.8M                              & Arabic, Cyrillic                    & The Scraping-Bys (1)                       & 188.6K                                \\
Chuvash                           & 1.5M                              & Cyrillic                            & The Scraping-Bys (1)                       & 191.0K                                \\
Turkmen                           & 921.0K                            & Arabic, Cyrillic, Latin             & The Scraping-Bys (1)                       & 191.7K                                \\
Bashkir                           & 893.1K                            & Cyrillic                            & The Scraping-Bys (1)                       & 713.9K                                \\
Uyghur                            & 343.0K                            & Arabic, Cyrillic, Latin             & The Scraping-Bys (1)                       & 187.0K                                \\
Karakalpak                        & 253.8K                            & Cyrillic, Latin                     & The Scraping-Bys (1)                       & {\color[HTML]{222222} 274.3K}         \\
Khakas                            & 219.0K                            & Cyrillic                            & The Left-Behinds (0)                       & 242.8K                                \\
Altai                             & 192.6K                            & Cyrillic                            & The Left-Behinds (0)                       & 190.0K                                \\
Crimean Tatar                     & 185.3K                            & Cyrillic, Latin                     & The Scraping-Bys (1)                       & 197.6K                                \\
Kumyk                             & 165.6K                            & Cyrillic                            & The Left-Behinds (0)                       & {\color[HTML]{222222} 192.4K}         \\
Karachay-Balkar                   & 162.8K                            & Cyrillic, Latin                     & The Scraping-Bys (1)                       & 182.6K                                \\
Gagauz                            & 157.4K                            & Cyrillic, Latin                     & The Scraping-Bys (1)                       & {\color[HTML]{222222} 177.1K}         \\
Sakha                             & 157.1K                            & Cyrillic                            & The Scraping-Bys (1)                       & 174.8K                                \\
Tuvinian                          & 103.2K                            & Cyrillic                            & The Scraping-Bys (1)                       & 148.3K                                \\
Shor                              & 2.3K                              & Cyrillic                            & The Left-Behinds (0)                       & {\color[HTML]{222222} 6.9K}           \\
Salar                             & 766                               & Latin                               & The Left-Behinds (0)                       & {\color[HTML]{222222} 1.5K}           \\
Urum                              & 491                               & Greek, Cyrillic, Latin              & The Left-Behinds (0)                       & {\color[HTML]{222222} 491}            \\ \hline
\end{tabular}
\caption{Corpus details for each Turkic language. \textit{Data} shows the aggregated amount of sentences across the corpus. \textit{Category} refers to the language classes based on data resource according to \cite{joshi2020state}}
\label{tab:data}

\end{table*}

%% file: bilingual.tex
\begin{table*}[t]
\centering
\resizebox{0.9\textwidth}{!}{%
\begin{tabular}{lrrrrrrrrrr}
\toprule
\textbf{Pair} &
  \textbf{Train size} &
  \textbf{Test size} &
  \multicolumn{2}{c}{\textbf{Bible}} &
  \textbf{Test size} &
  \multicolumn{2}{c}{\textbf{Ted Talks}} &
  \textbf{Test size} &
  \multicolumn{2}{c}{\textbf{X-WMT}}  \\ \cline{4-5} \cline{7-8} \cline{10-11}
 &
   &
   &
  \textbf{BLEU} &
  \textbf{ChrF} &
  \textbf{} &
  \textbf{BLEU} &
  \textbf{ChrF} &
   &
  \textbf{BLEU} &
  \textbf{ChrF} \\ \midrule

en-tr & 39.9m & 416 & 7.15 & 0.30 & \multicolumn{1}{r}{5.2k} & 12.32 & 0.43 & \multicolumn{1}{r}{800} & 19.87 & 0.51 \\ 
ru-tr & 16.8m & 455 & 7.44 & 0.33 & \multicolumn{1}{r}{5.1k} & 8.64  & 0.38  & \multicolumn{1}{r}{800} & 8.81 & 0.41  \\ \midrule

ru-uz & 1.22M & 684 & 6.01  & 0.41 & 2.7K & 4.51  & 0.76 & 800 & 5.95 & 0.39  \\ 
uz-ru & 1.22M & 684 & 9.84  & 0.51 & 2.7K & 7.57  & 0.73 & 800 & 7.45 & 0.37 \\ 
en-az & 784K  & 455 & 10.56 & 0.24 & 3.3K & 10.58 & 0.29 & 600 & 8.88 & 0.41  \\ 
az-en & 784K  & 455 & 21.17 & 0.45 & 3.3K & 17.01 & 0.17 & 600 & 12.14 & 0.42 \\ 
en-ky & 733K  & 451 & 6.47  & 0.32 & -    & -     & -    & 500 & 3.18  & 0.19 \\ 
ky-en & 733K  & 451 & 13.08  &  0.43 & -    & -     & -  & 500 & 4.30  & 0.40 \\ 
tr-az & 634K  & 606 & 13.78 & 0.65 & 3.6K & 20.50  & 0.40 & 500 & 9.68  & 0.33 \\ 
az-tr & 634K  & 606 & 11.66 & 0.71 & 3.6K & 24.20  & 0.95 & 500 & 11.53 & 0.49 \\ 
en-kk & 601K  & 453 & 3.62  & 0.61 & 3.6K & 6.31   & 0.29 & 700 & 6.99  & 0.38 \\ 
kk-en & 601K  & 453 & 11.22 & 0.27 & 3.6K & 9.78   & 0.30 & 700 & 9.75  & 0.46 \\ 
en-uz & 555K  & 465 & 5.23  & 0.40  & 3.2K & 5.89  & 0.20 & 800 & 6.60  & 0.42 \\ 
uz-en & 555K  & 465 & 16.20  & 0.63 & 3.2K & 11.61 & 0.18 & 800 & 12.32 & 0.48 \\ 
tr-uz & 161K  & 486 & 6.50   & 0.14 & 2.9K & 4.28  & 0.20 & 700 & 1.58  & 0.23  \\ 
uz-tr & 161K  & 486 & 7.40   & 0.32 & 2.9K & 3.92  & 0.26 & 700 & 1.73  & 0.22 \\ \midrule

kk-ky & 6.4K & 696 & 2.39 & 0.33 & - & - & - & 500 & 0.14 & 0.09  \\ 
ky-kk & 6.4K & 696 & 2.53 & 0.24 & - & - & - & 500 & 0.11   & 0.13  \\   
en-krc & 6.5K & 374 & 5.57  & 0.25 & - & - & - & -   & -    & -    \\ 
krc-en & 6.5K & 374 & 11.57 & 0.22 & - & - & - & -   & -    & -    \\ 
kk-tt  & 7.7K & 678 & 4.13  & 0.22 & - & - & - & -   & -    & -    \\ 
tt-kk  & 7.7K & 678 & 3.75  & 0.17 & - & - & - & -   & -    & -    \\ 
ru-sah & 8K   & 759 & 2.48  & 0.27 & - & - & - & 300 & 0.08 & 0.20  \\ 
sah-ru & 8K   & 759 & 2.44  & 0.23 & - & - & - & 300 & 0.31 & 0.16 \\ 
uz-kaa & 8.9K & 772 & 9.90   & 0.71 & - & - &  - & 300    & 5.39      & 0.41      \\ 
kaa-uz & 8.9K & 772 & 9.58  & 0.60  & - & - & - & 300  &   5.24   &  0.44      \\ \bottomrule

\end{tabular}%
}
\caption{Bilingual baselines separated by high-res., mid-res., and low-res. pairs. (K: thousand, M: million.)}
\label{tab:comb-bilingual}
\end{table*}

\begin{table*}[t]
\centering
\resizebox{0.85\textwidth}{!}{%
\begin{tabular}{lrrrrrrrrr}
\toprule
\multirow{2}{*}{\textbf{Pair}} &
  \multirow{2}{*}{\textbf{Test size}} &
  \multicolumn{2}{r}{\textbf{Baseline}} &
  \multicolumn{2}{r}{\textbf{Google Translate}} & 
  \multicolumn{2}{r}{\textbf{Yandex Translate}} &
  \multicolumn{2}{r}{\textbf{Apertium}} \\ \cline{3-10}
   &
   &
  \textbf{BLEU} &
  \textbf{ChrF} &
  \textbf{BLEU} &
  \textbf{ChrF} &
  \textbf{BLEU} &
  \textbf{ChrF} & 
  \textbf{BLEU} &
  \textbf{ChrF} \\ \hline
  
en--tr & 800 & 19.87 & 0.51  & \textbf{69.24} & 0.83 & 40.03 & 0.69 & --  & --    \\ 
ru--tr & 800 & 8.81 & 0.41  & \textbf{24.79} & 0.54 & 16.64 & 0.44 & --   & --    \\ 
% Medium Resource
tr--uz & 700  & 1.58  & 0.23 & \textbf{27.2}5 & 0.60 & 6.58  & 0.42  & --  & --   \\ 
uz--tr & 700  & 1.73 & 0.22 & \textbf{28.03} & 0.58 & 5.58  & 0.38 & 4.31 & 0.33   \\ 
en--uz & 800  & 6.60 & 0.42 & \textbf{48.50} & 0.72 &  15.66  & 0.51  & -- & --    \\ 
uz--en & 800  & 12.32 & 0.48 & \textbf{32.35} & 0.39 & 6.93  & 0.41 & -- & --  \\
en--kk & 700  & 6.99 & 0.38 & \textbf{26.60} & 0.55 & 5.51  & 0.39 & -- & --   \\ 
kk--en & 700  & 9.75 & 0.46 & 22.50 & 0.47 & \textbf{23.2}  & 0.50  & -- & --    \\ 
tr--az & 500  & 9.68 & 0.33 & \textbf{36.78} & 0.65 & 5.53  & 0.38  & -- & --   \\
az--tr & 500  & 11.53 & 0.49 & \textbf{32.67} & 0.62 & 11.75  & 0.44 & -- & --  \\ 
en--ky & 500  & 3.18 & 0.19  & \textbf{26.97} & 0.56 & 5.21 & 0.36    & --  & --  \\ 
ky--en & 500  & 4.30 & 0.40  & \textbf{21.66} & 0.50 & 3.89 & 0.20   & -- & --  \\ 
en--az & 600  & 8.88 & 0.41 & \textbf{78.54} & 0.89 & 6.59 & 0.40  & -- & --  \\ 
az--en & 600  & 12.14 & 0.42 & \textbf{39.42} & 0.65 & 12.54 & 0.46  & --  & -- \\ 
ru--uz & 800 & 5.95 & 0.39  & \textbf{22.26} & 0.56 & 13.19  & 0.50 & -- & --  \\ 
uz--ru & 800 & 7.45 & 0.37  & \textbf{19.00} & 0.48 & 10.87  & 0.43 & -- & --   \\ 
% Low resource
% en--krc* & 374 & 5.57  & 0.25 & -- & -- & -- & -- & --   & --    \\ \hline
% krc--en* & 374 & 11.57 & 0.22 & -- & -- & -- & -- & --   & --   \\ \hline
kk--tt*  & 678 & 4.13  & 0.22 & \textbf{5.45} & 0.35 & 1.58 & 0.24 & 2.77  & 0.28    \\ 
tt--kk*  & 678 & 3.75  & 0.17 & \textbf{5.44} & 0.35 & 1.41 & 0.22 & --   & --    \\ 
ru--sah & 300   & 0.08 & 0.20  & -- & -- & \textbf{8.27} & 0.40 & -- & --   \\ 
sah--ru & 300   & 0.31 & 0.16  & -- & -- & \textbf{24.93} & 0.54 & -- & --  \\
uz--kaa & 300 & 5.39   & 0.41 & -- & -- & -- &   &  \textbf{11.71}  & 0.42 \\ 
kaa-uz & 300 & \textbf{5.24} & 0.44 & -- & -- & -- & -- & 5.22 & 0.30 \\
kk-ky & 500 & 0.14 & 0.09 & \textbf{20.56} & 0.51 & 4.78 & 0.35 & 9.12 & 0.35 \\
ky-kk & 500 & 0.11 & 0.13 & \textbf{20.57} & 0.52 & 3.52 & 0.34 & 6.55 & 0.34 \\
\bottomrule

\end{tabular}%
}
\caption{Bilingual Baseline Compared to online MT Systems on X-WMT (Pair with * uses Bible Data)}
\label{tab:bilingual--hr--comparisons}
\end{table*}

%% file: datatable.tex
\begin{table*}[hbt!]
\centering
\begin{adjustwidth}{-0.8cm}{-0.8cm} 
\resizebox{17.5cm}{!}{%
\begin{tabular}{|l|r|r|r|r|r|r|r|r|r|r|r|r|r|r|r|r|r|r|r|r|l|r|l|r|}
\hline
 &
  \multicolumn{1}{c|}{\textbf{alt}} &
  \multicolumn{1}{c|}{\textbf{az}} &
  \multicolumn{1}{c|}{\textbf{ba}} &
  \multicolumn{1}{c|}{\textbf{cjs}} &
  \multicolumn{1}{c|}{\textbf{crh}} &
  \multicolumn{1}{c|}{\textbf{cv}} &
  \multicolumn{1}{c|}{\textbf{en}} &
  \multicolumn{1}{c|}{\textbf{gag}} &
  \multicolumn{1}{c|}{\textbf{kaa}} &
  \multicolumn{1}{c|}{\textbf{kjh}} &
  \multicolumn{1}{c|}{\textbf{kk}} &
  \multicolumn{1}{c|}{\textbf{krc}} &
  \multicolumn{1}{c|}{\textbf{kum}} &
  \multicolumn{1}{c|}{\textbf{ky}} &
  \multicolumn{1}{c|}{\textbf{ru}} &
  \multicolumn{1}{c|}{\textbf{sah}} &
  \multicolumn{1}{c|}{\textbf{slr}} &
  \multicolumn{1}{c|}{\textbf{tk}} &
  \multicolumn{1}{c|}{\textbf{tr}} &
  \multicolumn{1}{c|}{\textbf{tt}} &
  \multicolumn{1}{c|}{\textbf{tyv}} &
  \multicolumn{1}{c|}{\textbf{ug}} &
  \multicolumn{1}{c|}{\textbf{uum}} &
  \multicolumn{1}{c|}{\textbf{uz}} \\ \hline
\textbf{alt} &
  \multicolumn{1}{l|}{} &
  9.8K &
  11.8K &
  48 &
  6.8K &
  9.9K &
  11.3K &
  6.6K &
  6.8K &
  6.8K &
  6.8K &
  6.8K &
  6.7K &
  11.2K &
  6.8K &
  6.6K &
  61 &
  9.7K &
  11.3K &
  11.3K &
  \multicolumn{1}{r|}{4.5K} &
  6.8K &
   &
  10.5K \\ \hline
\textbf{az} &
  9.8K &
  \multicolumn{1}{l|}{} &
  33.0K &
  48 &
  6.9K &
  82.3K &
  787.6K &
  6.7K &
  6.9K &
  6.9K &
  8.0K &
  6.9K &
  6.7K &
  220.1K &
  389.0K &
  6.9K &
  52 &
  123.4K &
  636.5K &
  215.1K &
  \multicolumn{1}{r|}{13.6K} &
  6.9K &
   &
  227.4K \\ \hline
\textbf{ba} &
  11.8K &
  33.0K &
  \multicolumn{1}{l|}{} &
  \multicolumn{1}{l|}{} &
  6.8K &
  34.4K &
  64.5K &
  6.6K &
  6.8K &
  6.7K &
  6.8K &
  6.8K &
  6.7K &
  35.9K &
  36.2K &
  6.6K &
  \multicolumn{1}{l|}{} &
  30.2K &
  65.4K &
  67.2K &
  \multicolumn{1}{r|}{16.5K} &
  6.7K &
   &
  27.5K \\ \hline
\textbf{cjs} &
  48 &
  48 &
  \multicolumn{1}{l|}{} &
  \multicolumn{1}{l|}{} &
  49 &
  \multicolumn{1}{l|}{} &
  57 &
  48 &
  31 &
  47 &
  46 &
  \multicolumn{1}{l|}{} &
  \multicolumn{1}{l|}{} &
  47 &
  2.4K &
  48 &
  44 &
  47 &
  50 &
  50 &
   &
  53 &
   &
  48 \\ \hline
\textbf{crh} &
  6.8K &
  6.9K &
  6.8K &
  49 &
  \multicolumn{1}{l|}{} &
  6.7K &
  16.4K &
  6.8K &
  6.9K &
  7.0K &
  6.9K &
  6.9K &
  6.7K &
  6.7K &
  12.3K &
  6.7K &
  52 &
  6.7K &
  13.9K &
  8.3K &
   &
  7.0K &
   &
  7.3K \\ \hline
\textbf{cv} &
  9.9K &
  82.3K &
  34.4K &
  \multicolumn{1}{l|}{} &
  6.7K &
  \multicolumn{1}{l|}{} &
  156.4K &
  6.5K &
  6.7K &
  6.7K &
  6.8K &
  6.8K &
  6.6K &
  83.1K &
  82.9K &
  6.7K &
  \multicolumn{1}{l|}{} &
  78.6K &
  158.2K &
  160.0K &
  \multicolumn{1}{r|}{15.1K} &
  6.6K &
   &
  59.9K \\ \hline
\textbf{en} &
  11.3K &
  787.6K &
  64.5K &
  57 &
  16.4K &
  156.4K &
  \multicolumn{1}{l|}{} &
  6.8K &
  7.3K &
  7.0K &
  614.5K &
  7.0K &
  6.8K &
  625.6K &
  47.1M &
  7.1K &
  837 &
  250.4K &
  39.9M &
  572.5K &
  \multicolumn{1}{r|}{29.0K} &
  111.2K &
  \multicolumn{1}{r|}{507} &
  559.7K \\ \hline
\textbf{gag} &
  6.6K &
  6.7K &
  6.6K &
  48 &
  6.8K &
  6.5K &
  6.8K &
  \multicolumn{1}{l|}{} &
  6.7K &
  6.8K &
  6.7K &
  6.7K &
  6.6K &
  6.6K &
  6.7K &
  6.6K &
  55 &
  6.5K &
  6.7K &
  6.6K &
   &
  6.8K &
   &
  7.1K \\ \hline
\textbf{kaa} &
  6.8K &
  6.9K &
  6.8K &
  31 &
  6.9K &
  6.7K &
  7.3K &
  6.7K &
  \multicolumn{1}{l|}{} &
  6.8K &
  6.9K &
  6.9K &
  6.7K &
  6.7K &
  7.3K &
  6.7K &
  32 &
  6.6K &
  6.9K &
  6.8K &
   &
  6.8K &
   &
  9.8K \\ \hline
\textbf{kjh} &
  6.8K &
  6.9K &
  6.7K &
  47 &
  7.0K &
  6.7K &
  7.0K &
  6.8K &
  6.8K &
  \multicolumn{1}{l|}{} &
  6.8K &
  7.7K &
  6.7K &
  6.7K &
  6.8K &
  8.1K &
  52 &
  6.6K &
  6.8K &
  6.7K &
   &
  7.0K &
   &
  7.3K \\ \hline
\textbf{kk} &
  6.8K &
  8.0K &
  6.8K &
  46 &
  6.9K &
  6.8K &
  614.5K &
  6.7K &
  6.9K &
  6.8K &
  \multicolumn{1}{l|}{} &
  6.9K &
  6.7K &
  7.0K &
  4.5M &
  7.0K &
  48 &
  6.6K &
  65.9K &
  8.4K &
   &
  6.8K &
   &
  124.9K \\ \hline
\textbf{krc} &
  6.8K &
  6.9K &
  6.8K &
  \multicolumn{1}{l|}{} &
  6.9K &
  6.8K &
  7.0K &
  6.7K &
  6.9K &
  7.7K &
  6.9K &
  \multicolumn{1}{l|}{} &
  6.8K &
  6.7K &
  6.9K &
  7.6K &
  \multicolumn{1}{l|}{} &
  6.6K &
  6.9K &
  6.8K &
   &
  6.8K &
   &
  7.3K \\ \hline
\textbf{kum} &
  6.7K &
  6.7K &
  6.7K &
  \multicolumn{1}{l|}{} &
  6.7K &
  6.6K &
  6.8K &
  6.6K &
  6.7K &
  6.7K &
  6.7K &
  6.8K &
  \multicolumn{1}{l|}{} &
  6.6K &
  6.8K &
  6.6K &
  \multicolumn{1}{l|}{} &
  6.5K &
  6.7K &
  6.6K &
   &
  6.7K &
   &
  7.1K \\ \hline
\textbf{ky} &
  11.2K &
  220.1K &
  35.9K &
  47 &
  6.7K &
  83.1K &
  625.6K &
  6.6K &
  6.7K &
  6.7K &
  7.0K &
  6.7K &
  6.6K &
  \multicolumn{1}{l|}{} &
  309.3K &
  6.7K &
  50 &
  122.5K &
  549.0K &
  232.4K &
  \multicolumn{1}{r|}{15.0K} &
  6.7K &
   &
  127.0K \\ \hline
\textbf{ru} &
  6.8K &
  389.0K &
  36.2K &
  2.4K &
  12.3K &
  82.9K &
  47.1M &
  6.7K &
  7.3K &
  6.8K &
  4.5M &
  6.9K &
  6.8K &
  309.3K &
  \multicolumn{1}{l|}{} &
  8.7K &
  \multicolumn{1}{l|}{} &
  122.5K &
  16.8M &
  296.7K &
   &
  54.6K &
   &
  1.2M \\ \hline
\textbf{sah} &
  6.6K &
  6.9K &
  6.6K &
  48 &
  6.7K &
  6.7K &
  7.1K &
  6.6K &
  6.7K &
  8.1K &
  7.0K &
  7.6K &
  6.6K &
  6.7K &
  8.7K &
  \multicolumn{1}{l|}{} &
  51 &
  6.5K &
  6.9K &
  6.9K &
   &
  6.6K &
   &
  7.3K \\ \hline
\textbf{slr} &
  61 &
  52 &
  \multicolumn{1}{l|}{} &
  44 &
  52 &
  \multicolumn{1}{l|}{} &
  837 &
  55 &
  32 &
  52 &
  48 &
  \multicolumn{1}{l|}{} &
  \multicolumn{1}{l|}{} &
  50 &
  \multicolumn{1}{l|}{} &
  51 &
  \multicolumn{1}{l|}{} &
  56 &
  59 &
  32 &
   &
  59 &
   &
  48 \\ \hline
\textbf{tk} &
  9.7K &
  123.4K &
  30.2K &
  47 &
  6.7K &
  78.6K &
  250.4K &
  6.5K &
  6.6K &
  6.6K &
  6.6K &
  6.6K &
  6.5K &
  122.5K &
  122.5K &
  6.5K &
  56 &
  \multicolumn{1}{l|}{} &
  244.1K &
  235.4K &
  \multicolumn{1}{r|}{14.0K} &
  6.6K &
   &
  7.2K \\ \hline
\textbf{tr} &
  11.3K &
  636.5K &
  65.4K &
  50 &
  13.9K &
  158.2K &
  39.9M &
  6.7K &
  6.9K &
  6.8K &
  65.9K &
  6.9K &
  6.7K &
  549.0K &
  16.8M &
  6.9K &
  59 &
  244.1K &
  \multicolumn{1}{l|}{} &
  531.2K &
  \multicolumn{1}{r|}{29.2K} &
  64.4K &
   &
  254.8K \\ \hline
\textbf{tt} &
  11.3K &
  215.1K &
  67.2K &
  50 &
  8.3K &
  160.0K &
  572.5K &
  6.6K &
  6.8K &
  6.7K &
  8.4K &
  6.8K &
  6.6K &
  232.4K &
  296.7K &
  6.9K &
  32 &
  235.4K &
  531.2K &
  \multicolumn{1}{l|}{} &
  \multicolumn{1}{r|}{15.0K} &
  6.7K &
   &
  136.3K \\ \hline
\textbf{tyv} &
  4.5K &
  13.6K &
  16.5K &
  \multicolumn{1}{l|}{} &
  \multicolumn{1}{l|}{} &
  15.1K &
  29.0K &
  \multicolumn{1}{l|}{} &
  \multicolumn{1}{l|}{} &
  \multicolumn{1}{l|}{} &
  \multicolumn{1}{l|}{} &
  \multicolumn{1}{l|}{} &
  \multicolumn{1}{l|}{} &
  15.0K &
  \multicolumn{1}{l|}{} &
  \multicolumn{1}{l|}{} &
  \multicolumn{1}{l|}{} &
  14.0K &
  29.2K &
  15.0K &
   &
  \multicolumn{1}{l|}{} &
   &
  10.8K \\ \hline
\textbf{ug} &
  6.8K &
  6.9K &
  6.7K &
  53 &
  7.0K &
  6.6K &
  111.2K &
  6.8K &
  6.8K &
  7.0K &
  6.8K &
  6.8K &
  6.7K &
  6.7K &
  54.6K &
  6.6K &
  59 &
  6.6K &
  64.4K &
  6.7K &
   &
  \multicolumn{1}{l|}{} &
   &
  19.5K \\ \hline
\textbf{uum} &
  \multicolumn{1}{l|}{} &
  \multicolumn{1}{l|}{} &
  \multicolumn{1}{l|}{} &
  \multicolumn{1}{l|}{} &
  \multicolumn{1}{l|}{} &
  \multicolumn{1}{l|}{} &
  507 &
  \multicolumn{1}{l|}{} &
  \multicolumn{1}{l|}{} &
  \multicolumn{1}{l|}{} &
  \multicolumn{1}{l|}{} &
  \multicolumn{1}{l|}{} &
  \multicolumn{1}{l|}{} &
  \multicolumn{1}{l|}{} &
  \multicolumn{1}{l|}{} &
  \multicolumn{1}{l|}{} &
  \multicolumn{1}{l|}{} &
  \multicolumn{1}{l|}{} &
  \multicolumn{1}{l|}{} &
  \multicolumn{1}{l|}{} &
   &
  \multicolumn{1}{l|}{} &
   &
  \multicolumn{1}{l|}{} \\ \hline
\textbf{uz} &
  10.5K &
  227.4K &
  27.5K &
  48 &
  7.3K &
  59.9K &
  559.7K &
  7.1K &
  9.8K &
  7.3K &
  124.9K &
  7.3K &
  7.1K &
  127.0K &
  1.2M &
  7.3K &
  48 &
  7.2K &
  254.8K &
  136.3K &
  \multicolumn{1}{r|}{10.8K} &
  19.5K &
   &
  \multicolumn{1}{l|}{} \\ \hline
\end{tabular}%
}
\end{adjustwidth}
\caption{Parallel corpora size for each language pair}
\label{tab:til-data}
\end{table*}

%% file: sources.tex
\begin{table*}[t]
\centering
\begin{tabular}{ | p{0.3\linewidth}| p{0.3\linewidth} |  p{0.2\linewidth} | p{0.2\linewidth} |}
\hline
\textbf{Source}                                 & \textbf{Link}                                                                                  & \textbf{Languages}                                                                 & \textbf{Size}           \\
\hline
Tatoeba Challenge\newline(OPUS+Tatoeba+\newline Gourmet+JW300)  & \url{ https://github.com/Helsinki-NLP/Tatoeba-Challenge}                                              & az, ba, crh, cv, gag, kjh, kk, krc, kum, ky, sah, tk, tr, tt, tyv, ug, uz, ru, en                                                                        & $\sim$40m               \\ \hline
UDHR                                            & \url{https://www.ohchr.org/EN/UDHR/Pages/SearchByLang.aspx}                                          & alt, ba, az, cv, cjs, crh, gag, kaa, kjh, kk, ky, sah, slr, tk, tt, tr, ug, uz, ru, en& $\sim$100 per direction \\ \hline
Bible                                           & \url{https://www.faithcomesbyhearing.com/audio-bible-resources/recordings-database}                  & alt, ba, az, cjs,cv, crh, en, gag, kaa, kjh, kk, ky, sah, tk, tt, ug, uz, tr       & $\sim$9k per direction  \\ \hline
Ted Talks                                       & \url{https://www.ted.com/participate/translate/our-languages}                                        & az, en, kk, ky, ru, tt, tr, tt, uz, ug                                             & $\sim$600k              \\ \hline
Mozilla                                         &                                                                                                & az, ba, cv, en, kk, ky, sah, tk, tt, ug, uz, tr, ru                                & $\sim$300 per direction \\ \hline
Azerbaijani News                                & \url{https://github.com/derintelligence/en-az-parallel-corpus}                                     & az, en                                                                             & $\sim$68k               \\ \hline
Uzbek/English News                              & \url{https://data.gov.uz}\newline \url{https://president.uz}\newline \url{https://uz.usembassy.gov}\newline \url{https://www.gov.uz} & uz, en                                                                             & $\sim$60k               \\ \hline
Uzbekistan Legislative \newline Dataset (Law)            & \url{https://lex.uz/} & uz, ru, en                                                                         & $\sim$1.5m              \\\hline
KhanAcademy Project \newline Translations(Math/Science) & \url{https://uz.khanacademy.org/}                                                                    & uz, en                                                                             & $\sim$200k              \\\hline
Karakalpak News                                 & \url{https://kknews.uz} \newline
\url{https://www.gov.uz} \newline \url{http://karakalpakstan.uz} \newline \url{https://www.qrstat.uz/kk}  & kaa, uz, ru, en                                                                    & $\sim$60k               \\\hline
Bashkir-Russian Corpus                          & \url{https://github.com/AigizK/bashkort-parallel-corpora}                                            & ba,ru                                                                              & $\sim$600k              \\\hline
Salar Language Materials                        & \url{http://www.sino-platonic.org/complete/spp043\_salar\_language.pdf}                              & slr,en                                                                             & $\sim$700               \\\hline
Urum Language Materials                         & \url{https://web.archive.org/web/20180919233848/http://projects.turkmas.uoa.gr/urum/}                & urum, en                                                                           & $\sim$500               \\\hline
Russian-Shor \newline Online Dictionary                  & \url{http://tili.tadarlar.ru/tadar/rus-shor.html}                                                    & ru,cjs                                                                             & $\sim$300 \\
\hline
\end{tabular}
\caption{Sources and links for resources and websites used}
\label{tab:sources}
\end{table*}